\documentclass{article}

\usepackage[nonatbib,final]{nips_2018}

\usepackage[utf8]{inputenc} 
\usepackage[T1]{fontenc}    
\usepackage{hyperref}       
\usepackage{url}            
\usepackage{booktabs}       
\usepackage{amsfonts}       
\usepackage{nicefrac}       
\usepackage{microtype}      
\usepackage{graphicx}
\usepackage{amsmath}
\usepackage{amssymb}
\usepackage{bbm}
\usepackage{algorithm}
\usepackage{algorithmic}
\usepackage{color}
\usepackage{wasysym}

\title{Meta-Learner with Linear Nulling}

\author{
Sung Whan Yoon\\
\texttt{shyoon8@kaist.ac.kr}\\
\And
Jun Seo\\
\texttt{tjwns0630@kaist.ac.kr}\\
\And
Jaekyun Moon\\
\texttt{jmoon@kaist.edu}\\
\AND
\textnormal{School of Electrical Engineering,}\\
Korea Advanced Institute of Science and Technology (KAIST)
}

\begin{document}

\maketitle

\begin{abstract}
We propose a meta-learning algorithm utilizing a linear transformer that carries out null-space projection of neural network outputs. The main idea is to construct an alternative classification space such that the error signals during few-shot learning
 are quickly zero-forced on that space so that reliable classification on low data is possible. The final decision on a query is obtained utilizing a null-space-projected distance measure between the network output and reference vectors, both of which have been trained in the initial learning phase. Among the known methods with a given model size, our meta-learner achieves the best or near-best image classification accuracies with Omniglot and \textit{mini}ImageNet datasets.
\end{abstract}

\section{Introduction}

Achieving human-like adaptability remains a daunting challenge for developers of artificial intelligence.
In a related effort, researchers in the machine learning literature have tried to understand ways 
to learn quickly on a small number of training examples.
Learning the ability to quickly learn a new task on low data
is known as meta-learning, and typically two-level learning is employed: initial learning on 
large data sets representing widely varying tasks and few-shot learning on small amounts of unseen data 
to conduct a yet different task. Note that while the original meaning of meta-learning is to describe 
ability to conduct completely different sets of tasks without much new training in between \cite{SchmidhuberML, Schmidhuber93, Schmidhuber97, Thrun}, many researchers 
have also used the term meta-learning as they developed classification methods involving disjoint classes of images between initial training and inference.   

Significant advances have been made recently in this narrow sense of meta-learning, which is also often treated in the same context as few-shot learning. 
In particular, the prior work on combining a neural network with external memory,
known as the memory-augmented neural network (MANN) \cite{MANN}, showed a notable progress.
The Matching Network of \cite{MN} yields decisions based on matching the output of a network driven by a query sample to the output of another network fed by labeled samples; a training method was also introduced there which utilizes only a few examples per class at a time to match train and test conditions for a rapid learning.    
In yet another approach, the long short-term memory (LSTM) of \cite{LSTM} is trained to optimize another learner, which performs actual classification \cite{Ravi}. There, the parameters of the learner in few-shot learning are first set to the memory state of the LSTM, and then quickly learned based on the memory update rule of the LSTM, effectively capturing the knowledge from small samples. 
The model-agnostic meta-learner (MAML) of \cite{MAML} sets up the model for easy fine-tuning so that a small number of gradient-based updates allows the model to yield a good generalization.
A method dubbed the simple neural attentive meta-learner (SNAIL) combines an embedding neural network with temporal convolution and soft attention to draw from past experience 
\cite{SNAIL}.
The Prototypical Network of \cite{PN} makes decision by comparing the query output with the per-class 
cluster means in the embedding space while the network is learned via many rounds of 
episodic training.

In this paper, we propose a unique meta-learning algorithm utilizing linear transformation which allows classification 
in an alternative projection space to achieve improved few-shot learning.
During the initial meta-learning phase, a special set of vectors that acts as references for classification
is learned together with the embedding network. 
The linear transformer, which can be viewed as a simple null-space projector, zero-forces errors between
the per-class average outputs of the embedding network and the reference vectors in the projection space. 
The construction of this projection space is episode-specific.
The essence of our algorithm is
that when we try to match the network outputs representing the images to classify with the references utilized for classification, they do not need to be close in the original embedding space 
as long as they are conditioned to match well in the projection space. 
Our meta-learner exhibits competitive performances among existing meta-learners for Omniglot and \textit{mini}ImageNet image classification. In particular, in 20-way Omniglot experiments, our method gives near-best performance, second only to the SNAIL of \cite{SNAIL} in both 1-shot and 5-shot results. For 1-shot experiments of \textit{mini}ImageNet, our meta-learner is again the second to SNAIL, which yields the best result among all existing methods to our knowledge, except for the task-dependent adaptive metric (TADAM) of \cite{TADAM} that requires an extra network. 
For the 5-shot testing of \textit{mini}ImageNet, however, our method yields the best accuracy for a given model size, beating both the Prototypical Network of \cite{PN} and the SNAIL method of \cite{SNAIL}, albeit by a small margin against the latter.

\section{Meta-Learner with Linear Nulling} \label{section2}

\subsection{Model Description}

\begin{figure}
	\centering
	\includegraphics[width=120mm]{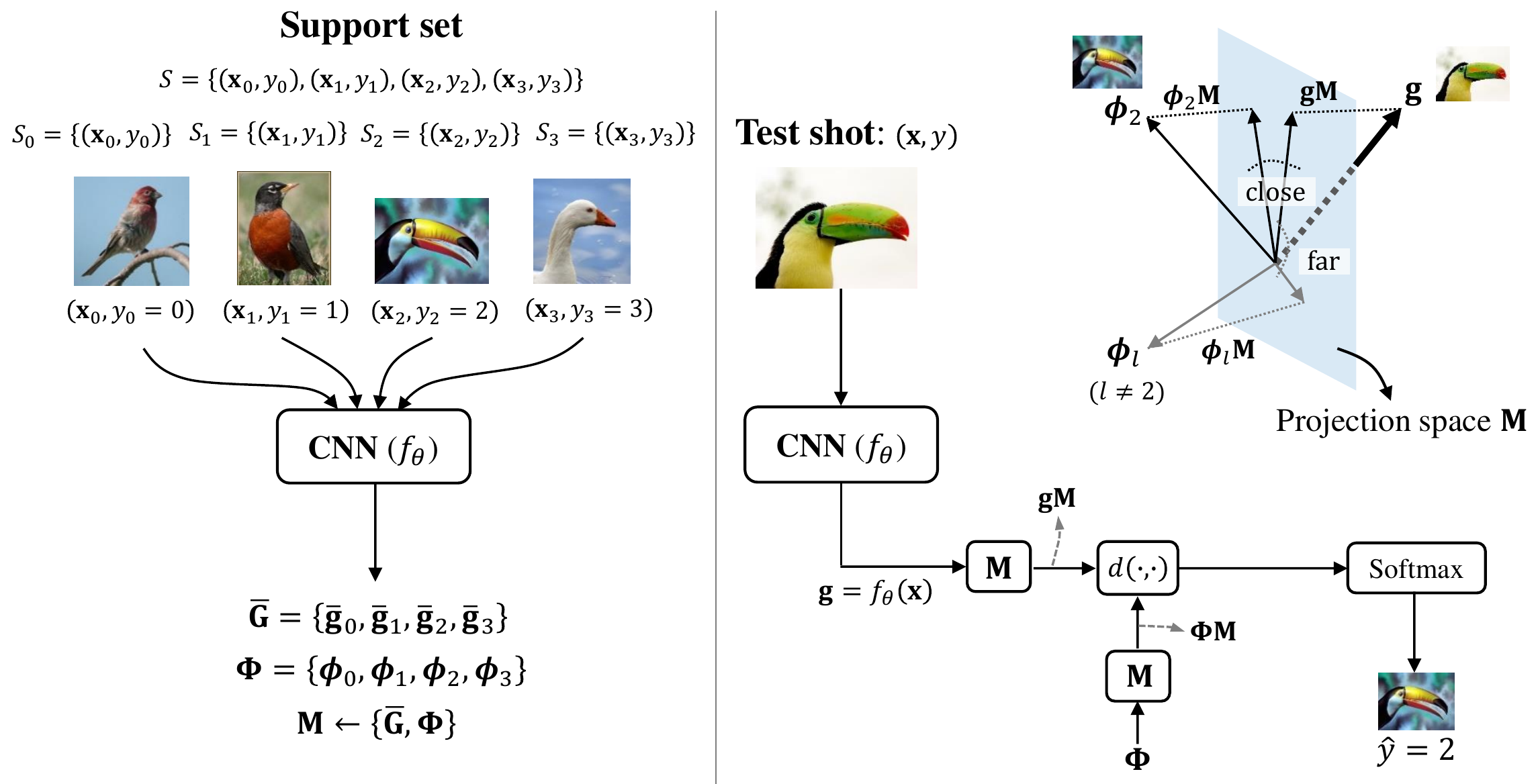}
	\caption{The proposed meta-learner with linear nulling}
	\label{fig:system_diagram}
\end{figure}

Let us provide a quick conceptual understanding of the proposed algorithm. In processing a given episode of labeled images (support set $\mathcal{S}$) and queries during the initial meta-learning phase, the convolutional neural network (CNN) output average $\bar{\mathbf{g}}_{k}$ is first generated for each label. A support set $\mathcal{S}$ contains pairs of images $\mathbf{x}_{n}$ and matching labels $y_n$. 
Let $S_{k}$ be the subset of $\mathcal{S}$ with a particular label $k$. See Fig. \ref{fig:system_diagram}. 
The model also makes use of a special set of vectors $\boldsymbol{\phi}_k$ that are carried over from the last episode-processing stage (arbitrarily initialized when handling the first episode). The purpose of $\boldsymbol{\phi}_k$'s will be made clear shortly.
Let $\mathbf{\Phi}$ be the matrix collecting $\boldsymbol{\phi}_k$'s for all labels. A projection space 
$\mathbf{M}$ is then constructed such that the average CNN output vectors $\bar{\mathbf{g}}_{k}$ and the vectors in $\mathbf{\Phi}$ are aligned closely in this space. 
The details of constructing $\mathbf{M}$ will be given below. 

Now, each query input is applied to the embedding CNN 
and the resulting output is collected. When this is done for all queries (with different labels),
both the CNN $f_{\theta}$ and the vectors in $\mathbf{\Phi}$ are adjusted by comparing the query output with 
the reference vectors such that
in-class similarities as well as out-of-class differences are maximized collectively in the projection space. 

This process gets repeated for every remaining episode with new classes of images and queries during the initial learning phase.
The vectors $\boldsymbol{\phi}_k$ act as references for classification, and they are learned from one episode to next together with the embedding network. Note that the projection space $\mathbf{M}$ itself is not learned but computed anew in every episode, given newly obtained averages $\bar{\mathbf{g}}_{k}$ and the reference vectors $\mathbf{\Phi}$ carried over from the last episode stage. The labels for the vectors in $\mathbf{\Phi}$ do not need to change from one episode to another. $\mathbf{M}$ can be seen as playing a crucial role in providing an efficient form of episode-specific conditioning. By the time the initial meta-learning phase is over, an effective few-shot learning strategy gets built-in jointly in the embedding network and the reference vector set.    

Now, as the few-shot learning phase begins, the learned parameters $f_{\theta}$ and $\mathbf{\Phi}$ will have all been fixed. First, a new set of per-class average vectors $\bar{\mathbf{g}}_{k}$ corresponding to the given few-shot images are obtained. Then, a new projection space is formed using these average vectors as well as the reference vectors in $\mathbf{\Phi}$. 
Next, the network output for the test shot is compared with each of the reference vectors in the projection space for the final classification decision. The Euclidean distance is used here as distance measure and softmax is utilized as the activation function. 

\subsection{Construction of Null-Space}

The projection space $\mathbf{M}$ is obtained as follows. Consider the training samples labeled $k$ in a given episode.
The average CNN outputs $\bar{\mathbf{g}}_{k}$ are obtained for all classes.
As mentioned, we are to choose $\mathbf{M}$ such that $\bar{\mathbf{g}}_{k}$
and the matching reference weight vector $\boldsymbol{\phi}_k$ are aligned closely when projected on $\mathbf{M}$ itself. 
At the same time, we also want to make $\bar{\mathbf{g}}_{k}$ and the non-matching weights $\boldsymbol{\phi}_l$ for all $l\neq k$ well-separated in the same projection space. An effective way to achieve this is to maximize
\begin{align} \label{eq:delta_rep_key}
\Delta_{k} = \big\{(N_{c}-1)\boldsymbol{\phi}_{k} - \sum_{{\scriptstyle l\neq k }}{\boldsymbol{\phi}_{l}} \big\} \mathbf{M}\mathbf{M}^{T} \bar{\mathbf{g}}_{k}^{T}
\end{align}
which is to say that two matching vectors $\boldsymbol{\phi}_{k}$ and $\bar{\mathbf{g}}_{k}$ 
yield a large inner product (i.e., well-aligned) on $\mathbf{M}$ while at the same time the non-matching vector sum $\sum_{l\neq k}{\boldsymbol{\phi}_{l}}$ and $\bar{\mathbf{g}}_{k}$ have a small inner product.
$N_c$ is the number of labels. The class references $\boldsymbol{\phi}_{k}$'s are not relabeled from one episode to next.

We assume that $\mathbf{M}$ takes the form of a matrix whose columns are orthonormal unit vectors.
Defining the difference or error vector $\mathbf{v}_{k} = \{(N_{c}-1)\boldsymbol{\phi}_{k} - \sum_{l\neq k}{\boldsymbol{\phi}_{l}} \} - \bar{\mathbf{g}}_{k}$, maximization of (1) is done by taking $\mathbf{M}$ to be a subspace of the orthogonal complement of the space spanned by $\mathbf{v}_{k}$ in $\mathbb{R}^{D}$, where $D$ is the length of $\mathbf{v}_{k}$.
In this way, the error vector is always forced to zero when projected on $\mathbf{M}$.
To consider all classes, $\mathbf{M}$ should be a subspace of the null-space of a matrix whose columns are $N_{c}$ error vectors, i.e., $\mathbf{M} \subseteq \text{null}\big( \{ \mathbf{v}_{0},\cdots,\mathbf{v}_{N_{c}-1} \} \big)$.

By focusing only on the null-space, we are essentially cutting out unnecessary subspaces for classification that contain non-orthogonal basis to any error vectors. 
With such an $\mathbf{M}$, all error vectors are forced to zero when projected on itself.
Note that a simple choice of $\mathbf{M} = \text{null}\big( \{ \mathbf{v}_{0},\cdots,\mathbf{v}_{N_{c}-1} \} \big)$
is a perfectly good solution, which we employ in this paper.
If $D$ is larger than $N_{c}$, the projection space $\mathbf{M}$ always exists.
Note that because the matching between the network output averages and the reference vectors occurs on the 
episode-specific projection space, $\boldsymbol{\phi}_{k}$'s do not need to be relabeled in every episode; the same label sticks to each reference vector throughout the initial learning phase.

\section {Experiment Results}\label{section3}

\subsection{Settings for Omniglot and \textit{mini}ImageNet Few-shot Classification}
Omniglot \cite{Omniglot} is a set of images of  1623 handwritten characters from 50 alphabets with 20 examples for each class. We have used 28$\times$28 downsized grayscale images and introduced class-level data augmentation by random angle rotation of images in multiples of 90$^\circ$ degrees, as done in prior works \cite{MANN,PN,MN}. 1200 characters are used for training and test is done with the remaining characters.
\textit{mini}ImageNet is a dataset suggested by Vinyals et al. for few-shot classification of colored images \cite{MN}. It is a subset of the ILSVRC-12 ImageNet dataset \cite{Imagenet} with 100 classes and 600 images per each class. We have used the splits introduced by Ravi and Larochelle \cite{Ravi}. For experiment, we have used 84$\times$84 downsized color images with a split of 64 training classes, 16 validation classes and 20 test classes. Data augmentation was not employed in the \textit{mini}ImageNet experiment. The Adam optimizer \cite{Adam} with optimized learning-rate decay is used.



For fair comparison, we employ the same embedding CNN widely used in prior works. It is based on four convolutional blocks, each of which consists of a 3$\times$3 2D convolution layer with 64 filters, stride 1 and padding, a batch normalization layer \cite{BN}, a ReLU activation and a 2$\times$2 maxpooling, as done in \cite{SNAIL,PN,MN}. 

For 20-way Omniglot and 5-way \textit{mini}ImageNet classification, our meta-learner is trained with 60-way and 20-way episodes, respectively. In the test phase of 20-way Omniglot and 5-way \textit{mini}ImageNet classification, we have to choose 20 and 5 reference vectors among 60 and 20, respectively. In selecting only a subset of reference vectors for testing purposes, relabeling is done. For each average network output chosen in arbitrary order, the closest vector among the remaining ones in $\mathbf{\Phi}$ is tagged with the matching label.
The closeness measure is the Euclidean distance in our experiments. 
After choosing the closest reference vectors, the projection space $\mathbf{M}$ is obtained for few-shot classification. The experimental results of our meta-learner in Table \ref{acc_table1} are based on 60-way initial learning for 20-way Omniglot and 20-way initial learning for 5-way \textit{mini}ImageNet classification.

\subsection{Results}

In Table \ref{acc_table1}, few-shot classification accuracies of our meta-learner with nulling (MLN) are presented. The performance in the 20-way Omniglot experiment is evaluated by the average accuracy over randomly chosen $1\times10^{4}$ test episodes with 5 query images for each class. On the other hand, the performance in 5-way \textit{mini}ImageNet is evaluated by the average accuracy and a 95\% confidence interval over randomly chosen $3\times10^{4}$ test episodes with 15 query images for each class.

When we obtain the projection space, we normalize $\boldsymbol{\Phi}$ and $\bar{\mathbf{g}}_{k}$ by forcing the power of all vectors to 1. Once the projection space is found, classification of the query is done by measuring the distance to the reference vectors without normalizing the vectors, since the query output is not normalized.



Prior meta-learners including the matching networks \cite{MN}, MAML \cite{MAML}, the Prototypical Networks \cite{PN} and SNAIL \cite{SNAIL} are compared.
For 20-way Omniglot classification, MLN shows the second best performance for both 1- and 5-shot cases. Although the performance is not the best, the classification accuracies are fairly close to the best.
For the 5-shot case of 5-way \textit{mini}ImageNet classification, our meta-learner achieves the best performance. 
We remark that for \textit{mini}ImageNet, the recently introduced TADAM of \cite{TADAM} actually shows the best performance 
among all known methods including our MLN. However, we opted not to include this method in the comparison table as it
requires an extra network for the task-conditioning and thus comparison would not be fair.

Our meta-learner is perhaps closest in spirit to the Prototypical Network, among known prior methods, in that
both methods rely on training references via repetitive episode-specific conditioning with the 
final query image compared against each of the references during classification. The key difference is that
in the Prototypical Network, the generalization strategy is directly learned in the embedding network,
whereas in our method a separate set of reference vectors is maintained which is learned together with the network. Employing a projection space to perform classification is also a unique feature of MLN.  

\begin{table}[h]
  \caption{Few-shot classification accuracies for 20-way Omniglot and 5-way \textit{mini}ImageNet}
  \label{acc_table1}
  \centering
  \begin{tabular}{l||cc||cc}
    \toprule  
    & \multicolumn{2}{c}{\textbf{20-way Omniglot}} & \multicolumn{2}{c}{\textbf{5-way \textit{mini}ImageNet}} \\
    \cmidrule{2-5}
    \textbf{Methods}    & 1-shot & 5-shot	& 1-shot & 5-shot  \\
    \midrule
    \textbf{Matching Nets} \cite{MN}  & 88.2\%	& 97.0\% & 43.56 $\pm$ 0.84\%  & 55.31 $\pm$ 0.73\%   \\
    \textbf{MAML} \cite{MAML} & 95.8\%	& 98.9\% & 48.70 $\pm$ 1.84\%  & 63.15 $\pm$ 0.91\%	 \\
    \textbf{Prototypical Nets} \cite{PN} & 96.0\%	& 98.9\% & 49.42 $\pm$ 0.78\%  & 68.20 $\pm$ 0.66\%	  \\
    \textbf{SNAIL} \cite{SNAIL}  & 97.64\%	& 99.36\% & 55.71 $\pm$ 0.99\%  & 68.88 $\pm$ 0.92\%  \\
    \midrule
    \textbf{MLN}    & 97.33\%	& 99.18\%  &50.68 $\pm$ 0.11\%  & \textbf{69.00} $\pm$ \textbf{0.09}\%\\
    \bottomrule
  \end{tabular}
\end{table}

\section{Conclusion}\label{section5}
In this work, we proposed a meta-learning algorithm aided by a linear transformer that 
performs null-space projection of the network output.
Our algorithm uses linear nulling
to shape the classification space where network outputs could be better classified. 
Our meta-learner achieves the best or near-best performance among known methods in various few-shot image classification tasks for a given network size.



\subsubsection*{Acknowledgments}
This work is supported by the ICT R\&D program of Institute for Information \& Communications
Technology Promotion (IITP) grant funded by the Korea government (MSIP) [2016-0-005630031001, Research on Adaptive Machine Learning Technology Development for Intelligent Autonomous Digital Companion].

\small

\bibliographystyle{plain}
\bibliography{NIPS2018}

\end{document}